\documentclass{article}

\usepackage[preprint]{neurips_2025}

\usepackage[utf8]{inputenc} 

\usepackage[T1]{fontenc}    
\usepackage{hyperref}       
\usepackage{url}            
\usepackage{booktabs}       
\usepackage{amsfonts}       
\usepackage{graphicx}       
\usepackage{nicefrac}       
\usepackage{microtype}      
\usepackage{xcolor}         
\usepackage{graphicx}       
\usepackage{algorithm}      
\usepackage{algorithmic}    
\usepackage{amsmath}        
\usepackage{multirow}
\usepackage{bbding}
\usepackage{colortbl}

\newcommand{\DriveMRP}[1]{{DriveMRP}}
\newcommand{\ourdata}[1]{{DriveMRP-10K}}
\newcommand{\ourmethod}[1]{{DriveMRP-Agent}}

\title{
DriveMRP: Enhancing Vision-Language Models with Synthetic Motion Data for \underline{M}otion \underline{R}isk \underline{P}rediction
}

\author{
    \textbf{Zhiyi Hou}\textsuperscript{1,2,3,*}, 
    \textbf{Enhui Ma}\textsuperscript{1,3,*}, 
    \textbf{Fang Li}\textsuperscript{2,*},
    \textbf{Zhiyi Lai}\textsuperscript{2},
    \textbf{Kalok Ho}\textsuperscript{2}, \\
    \textbf{Zhanqian Wu}\textsuperscript{2},
    \textbf{Lijun Zhou}\textsuperscript{2},
    \textbf{Long Chen}\textsuperscript{2},
    \textbf{Chitian Sun}\textsuperscript{2},
    \textbf{Haiyang Sun}\textsuperscript{2,$\dagger$}, \\
    \textbf{Bing Wang}\textsuperscript{2},
    \textbf{Guang Chen}\textsuperscript{2},
    \textbf{Hangjun Ye}\textsuperscript{2},
    \textbf{Kaicheng Yu}\textsuperscript{1,\Envelope} \\  
    \\
    \textsuperscript{1}Westlake University \hspace{1em}
    \textsuperscript{2}Xiaomi EV \hspace{1em}
    \textsuperscript{3}Zhejiang University \\
    {\tt\small houzhiyi@westlake.edu.cn} \\
}

\begin{document}
\maketitle

\let\thefootnote\relax
\footnotetext{
\small
\textsuperscript{*} Equal contribution, 
\textsuperscript{$\dagger$} Project lead,
\textsuperscript{\Envelope} Corresponding author.  
}

\begin{figure}[h!]
    \centering
    \includegraphics[width=\linewidth]{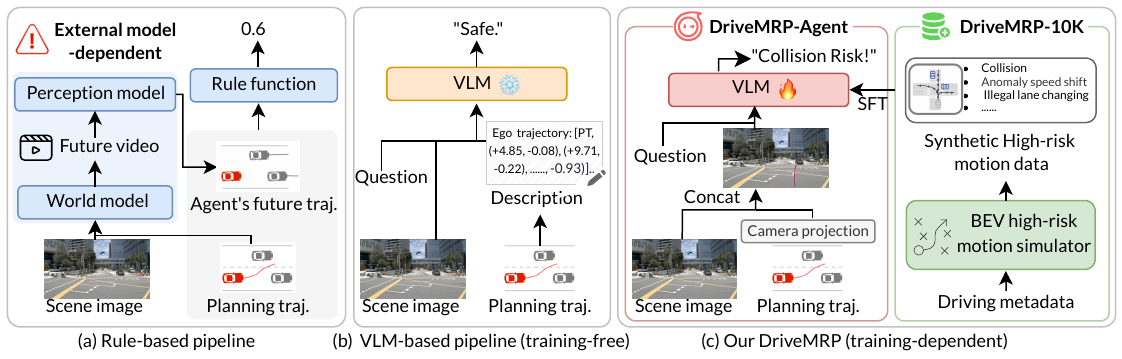}
    \vspace{-0.2cm}
    \caption{
      (a) Previous rule-based pipelines rely on external models to predict other vehicles' future positions, which are highly sensitive to perception error from these external models, and their output scores lack interpretability.
      (b) Recent VLM-based pipelines convert ego motion into discretized actions described in natural language for risk prediction, but they ignore the modality gap between language and vision, leading to suboptimal performance.
      (c) In this paper, we enhance VLM's motion risk prediction capability through SFT using synthetic data \textbf{\ourdata{}}. Moreover, we introduce a VLM-agnostic framework called \textbf{\ourmethod{}}, which employs a projection-based visual prompting scheme to address the modality gap issue. Experiments demonstrate that our synthetic-data-based approach can generalize zero-shot to the real world.
    }
    \label{fig: teaser}
    \end{figure}

\begin{abstract}
    
Autonomous driving has seen significant progress, driven by extensive real-world data. However, in long-tail scenarios, accurately predicting the safety of the ego vehicle's future motion remains a major challenge due to uncertainties in dynamic environments and limitations in data coverage. In this work, we aim to explore whether it is possible to enhance the motion risk prediction capabilities of Vision-Language Models (VLM) by synthesizing high-risk motion data. Specifically, we introduce a Bird's-Eye View (BEV) based motion simulation method to model risks from three aspects: the ego-vehicle, other vehicles, and the environment. This allows us to synthesize plug-and-play, high-risk motion data suitable for VLM training, which we call \ourdata{}. Furthermore, we design a VLM-agnostic motion risk estimation framework, named \ourmethod{}. This framework incorporates a novel information injection strategy for global context, ego-vehicle perspective, and trajectory projection, enabling VLMs to effectively reason about the spatial relationships between motion waypoints and the environment. Extensive experiments demonstrate that by fine-tuning with \ourdata{}, our \ourmethod{} framework can significantly improve the motion risk prediction performance of multiple VLM baselines, with the accident recognition accuracy soaring from 27.13\% to 88.03\%. Moreover, when tested via zero-shot evaluation on an in-house real-world high-risk motion dataset, \ourmethod{} achieves a significant performance leap, boosting the accuracy from base\_model's 29.42\% to 68.50\%, which showcases the strong generalization capabilities of our method in real-world scenarios.
More information can be found on our \href{https://github.com/xiaomi-research/drivemrp}{\textcolor{blue}{GitHub repository}}.

\end{abstract}


\section{Introduction}

End-to-end autonomous driving has recently witnessed rapid advancements~\cite{hu2023uniad, Jiang2023VADVS,wang2024omnidrive,sima2024drivelm,tian2024drivevlm}. However, accurately predicting the safety of an ego-vehicle's future motion in long-tail scenarios remains a significant challenge, primarily due to the inherent uncertainties of dynamic environments and the limitations in existing data coverage. In these complex situations, the trajectory planning module often generates multiple potential trajectories of varying quality. 
Existing trajectory evaluation methods typically use rule-based functions~\cite{dosovitskiy2017carla,li2022metadrive,dauner2024navsim,caesar2021nuplan} or learning-based models~\cite{wang2024Drive-WM,gao2024vista,huang2024Gen-Drive} to directly output a single reward score without any explanation of risk categories, which does not assist end-to-end algorithms in taking corresponding prevention measures. Therefore, we argue that identifying the types of risks and providing explanations for the causes of those risks are foundational for reliable autonomous driving and continuous improvement of decision-making algorithms.

The difficulties in robust motion risk prediction stem from two primary aspects. Firstly, high environmental uncertainty, particularly in comprehending unstructured spaces~\cite{min2024autonomousUnstructured,seo2023safe}  (e.g., weather~\cite{zhang2023perceptionweather}) and the unpredictable behavior of dynamic traffic participants~\cite{chen2024autonomous,zhou2024UA-Track} (e.g., other vehicles, pedestrians, cyclists). Secondly, data limitations persist. Learning-based methods are often constrained by the quantity and diversity of real-world data~\cite{li2024egostatus}, especially concerning rare, high-risk events. Insufficient coverage of such scenarios makes models susceptible to out-of-distribution states, hampering their generalization to unseen, dangerous situations.

As illustrated in Figure~\ref{fig: teaser}(a), due to the unknown intentions of other agents in the scene, traditional rule-based methods (e.g., Drive-WM) must rely on external world models and perception models to predict the future coordinate positions of other vehicles, and subsequently calculate scores based on pre-defined rules. This approach is highly sensitive to inaccuracies in the world model and perception, making it difficult to accurately assess future risks. Secondly, their heavy reliance on structured spaces hinders their generalization to the real world. As shown in Figure~\ref{fig: teaser}(b), recent works (e.g., OmniDrive, DriveLM) attempt to leverage the rich understanding of unstructured spaces by Vision-Language Models (VLMs) to identify high-risk motions. However, they perform direct zero-shot inference, using numerical text formats to input trajectory coordinates. The significant modality gap between numerical coordinates and visual information makes it difficult for VLMs to deeply understand the critical spatial relationships between motion waypoints and the surrounding environment, leading to suboptimal risk prediction results.

In this work, we explore the potential of enhancing VLM motion risk prediction capabilities by synthesizing scalable high-risk motion data, as demonstrated in Figure~\ref{fig: teaser}(c). Specifically, we introduce a Bird's-Eye View (BEV) based motion simulation method for risk modeling across three dimensions: the ego-vehicle, other vehicles, and the environment (e.g., potential collisions, abrupt velocity changes). This enables us to synthesize plug-and-play, high-risk motion data suitable for VLM training, e.g., \ourdata{}. Furthermore, we design a VLM-agnostic motion risk prediction framework, e.g., \ourmethod{}. This framework incorporates a motion projection-based visual prompting scheme, expressing motion in a visual form. This simple scheme addresses the modality gap between numerical waypoint and images, thereby enabling VLMs to effectively reason spatial relationships between motion waypoints and the environment. Then,  mimicking human-like chain-of-thought processes, we introduce a multi-step logical visual QA approach to guide the VLM in progressively reasoning about potential risks through two stages: scene understanding and motion analysis.
In summary, the main contributions are as follows:

\begin{itemize}
    \item  A BEV-based motion simulation pipeline capable of synthesizing scalable high-risk motion data, which is capable of augmenting VLM training for risk prediction enhancement.
    
    \item A synthetic high-risk motion dataset (\ourdata{}), and a benchmark focusing on challenging high-risk motions to facilitate research in VLM-based motion risk prediction.

    \item A VLM-agnostic motion risk prediction framework (\ourmethod{}), featuring a novel visual prompting scheme, enables comprehensive scene understanding and trajectory risk prediction in complex scenarios.

\end{itemize}

\section{Related Work}


\noindent\textbf{Rule-based Motion Reward Estimation.}
In autonomous driving, rule-based reward functions typically rely on simulators~\cite{dosovitskiy2017carla,caesar2021nuplan,li2022metadrive,dauner2024navsim} to evaluate vehicle motion dynamics and compliance. However, existing simulators provide limited types of structured scene annotations (e.g., map elements, traffic signal states), restricting the capacity of such methods for accurate and generalizable assessment in complex, real-world dynamic scenes. For instance, mainstream simulators~\cite{caesar2021nuplan,dauner2024navsim} lack precise annotations for map-level solid/dashed lines, real-time traffic light states, and variable weather conditions. Consequently, they struggle to accurately identify behaviors such as running red lights or illegal lane crossings, nor can they effectively evaluate risks associated with hazardous driving in extreme weather. This limitation makes it difficult for purely rule-based reward estimation to cover all critical risk factors in real-world driving.

\noindent\textbf{World Model-based Motion Reward Estimation.}
Recently, some research has explored using world models for reward estimation in motion planning. For example, Drive-WM~\cite{wang2024Drive-WM} employs action-conditioned video generation to predict future scenarios, subsequently using external perception modules~\cite{li2024bevformer,liao2022maptr} for structured scene representation and rule-based reward calculation. This approach inherits the limitations of rule-based methods aforementioned and its perception module, trained on specific datasets~\cite{caesar2020_nuscenes}, struggles with generalization. Another work, Vista~\cite{gao2024vista}, uses a video prediction model pre-trained on expert data to evaluate candidate actions. However, its inherent expert bias favors motions similar to expert trajectories, facing imitation learning's generalization challenges and hindering exploration of novel strategies. Furthermore, both approaches suffer from poor interpretability, outputting scalar rewards without clear explanations. Thus, we explore to leverage the interpretability of Vision-Language Models (VLMs) for more transparent risk prediction.

\noindent\textbf{VLM-based Motion Evaluation.}
Pre-trained VLMs show significant potential in end-to-end autonomous driving due to their strong open-world understanding and reasoning, becoming a research hotspot~\cite{EM-VLM4AD,ELM,tian2024drivevlm,sima2024drivelm,wang2024omnidrive,mao2023Gpt-driver,han2025DME-Driver}. Nevertheless, directly using VLMs for fine-grained motion risk prediction remains less explored. Existing attempts, like HE-Drive~\cite{wang2024HE-Drive}, use VLMs to adjust weights of traditional rule-based metrics but still heavily rely on hand-crafted rules. Another work, Gen-Drive~\cite{huang2024Gen-Drive}, uses VLMs to assist in creating trajectory preference datasets for training reward models. However, its data, primarily sampled from biased 
diffusion models, consists mostly of common safe scenarios, lacking coverage of dangerous or critical situations. Consequently, the resulting reward models have limited ability to evaluate rare, high-risk behaviors.
To address these challenges, we propose a Bird's-Eye View (BEV)-based simulation pipeline that models risk behavior to synthesize critical high-risk motion data, which is difficult to collect at scale in the real world. Empirical study proves that our synthesized dataset can significantly enhance the performance of general-purpose VLMs for motion risk prediction in autonomous driving.

\section{Method}

\subsection{Overview}

This paper introduces a novel task for autonomous driving motion risk prediction, with the primary objective of identifying the risk categories associated with the ego vehicle's future motion trajectories planned by end-to-end autonomous driving algorithms, and providing corresponding causal explanations. This initiative aims to furnish crucial insights for the continuous optimization and iteration of decision-making algorithms. We define the ego vehicle's future motion as $M$, which comprises a sequence of $N$ waypoints $(x,y)$ in a bird's-eye view (BEV) perspective. Formally, $M=\{(x_0,y_0),(x_1,y_1),\ldots,(x_N,y_N)\}$, where each waypoint indicates the ego vehicle's future position relative to its current position after a constant time step $n=(0,\ldots,N)$. The output of the motion risk prediction task encompasses three aspects: a comprehensive understanding of the driving scene, the identified potential risk categories, and an in-depth analysis of the risk etiology.

To achieve this, we first elaborate in Section~\ref{sec:ourdata} on a pipeline for efficiently synthesizing high-risk driving behavior data within a simulated environment. This pipeline is designed to address the current deficiency in authentic high-risk behavior evaluation benchmarks within the autonomous driving domain. The essence of our data synthesis process lies in the automated, large-scale generation of diverse trajectory data by precisely simulating high-risk behaviors atop existing raw autonomous driving data. In Section~\ref{sec:ourmetric}, we introduce an adapted evaluation metric tailored to assess the proposed task. Furthermore, as detailed in Section~\ref{sec:ourmethod}, leveraging this synthesized high-quality dataset, we propose a Vision-Language Model (VLM)-based baseline model, termed \ourmethod{}. This agent not only possesses robust scene understanding capabilities but can also accurately determine whether the future trajectories predicted by an autonomous driving system will lead to high-risk situations, based on the current complex driving environment.

\begin{figure}[t!]
    \centering
      \includegraphics[width=\linewidth]{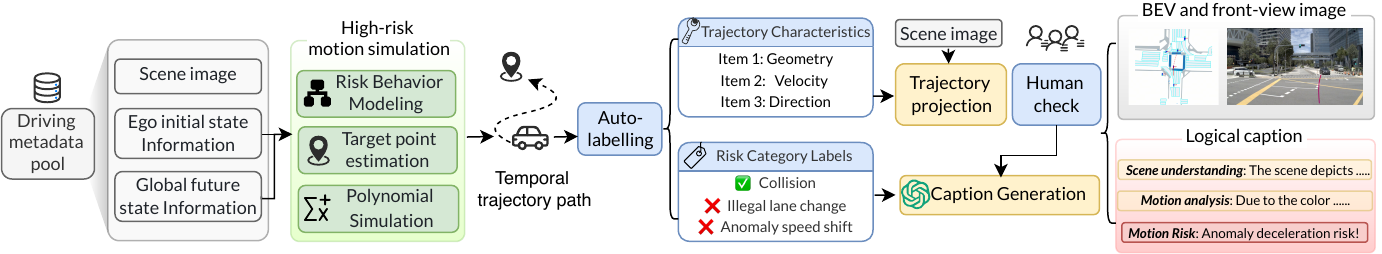}
    \vspace{-0.2cm}
    \caption{
    The proposed \ourdata{} simulation pipeline consists of the following stages: high-risk motion synthesis based on polynomial simulation, automatic labeling of motion attributes, quality check with human-in-the-loop, and caption generation.
    }
    \label{fig: ourdata} 
\end{figure}

\subsection{\ourdata{}}
\label{sec:ourdata}

We introduce \ourdata{}, a large-scale, synthetically generated dataset of high-risk motions built upon the nuPlan dataset. \ourdata{} consists of high-quality visual question answering (VQA) pairs, specifically tailored for the motion risk prediction task in autonomous driving. The detailed procedure for our data synthesis is presented below. 
The \ourdata{} simulation pipeline incorporates a "Human-in-the-Loop" (HITL) VQA pair generation mechanism, which integrates trajectory simulation techniques with the potent capabilities of the GPT-4o~\cite{achiam2023gpt} model. As illustrated in Figure 2, the entire data generation process can be distinctly segmented into the following core stages: high-risk trajectory synthesis based on polynomial simulation, automated annotation of motion attributes, rigorous human-in-the-loop quality control, and the final caption generation.

\subsubsection{BEV-based High-Risk Motion Simulation}

The simulation of high-risk motion is a pivotal component of our data synthesis strategy, where we systematically design and generate high-risk scenarios by considering three critical dimensions: ego-vehicle behavior, interactions with other vehicles, and environmental constraints.

\noindent\textbf{High-Risk Scenario Selection.} Based on a statistical analysis of a large corpus of real-world traffic accidents and high-risk driving behaviors, we selected one or more representative hazardous events for each of the aforementioned dimensions to simulate. Specifically, for ego-vehicle behavior, we focus on Hard Braking and Abnormal Acceleration scenarios to assess the precise understanding of the ego-vehicle's state. For interactions with other vehicles, we primarily simulate Collision events, which directly reflect the system's capability to predict the intentions of dynamic traffic participants (e.g., other vehicles, pedestrians) and identify interaction risks. Regarding environmental and map constraints, we selected Lane Violation / Off-Road Departure scenarios to test the system's understanding of and adherence to static environmental elements such as road markings and drivable areas. These meticulously chosen scenarios cover critical information and hazardous behavior patterns widely recognized in driving.

\noindent\textbf{Rule Definition for High-Risk Scenarios.} To ensure that the simulated high-risk scenarios are both realistic and possess clear evaluation criteria, we have formulated precise rule-based definitions for each selected high-risk scenario. For instance, collision events are determined by a predefined minimum safety distance threshold. Hard acceleration or deceleration is characterized by the magnitude and duration thresholds of acceleration or deceleration. Abnormal lane changes or line crossings are judged based on the geometric relationship between the vehicle's trajectory and lane markings or road edges. In defining these rules, we have extensively considered the actual vehicle dynamics, such as normal operating speed ranges, and the smoothness and continuity of trajectories, ensuring that the generated trajectories are physically plausible and logically coherent.

\noindent\textbf{Simulation of High-Risk Scenarios.} Within each raw scene segment provided by the nuPlan dataset~\cite{caesar2021nuplan}, we generate specific high-risk scenarios by leveraging the scene's initial state (e.g., vehicle position, velocity, acceleration) and the evolving environmental trends over a future time horizon, in conjunction with the predefined rule constraints. Operationally, we first deduce critical target points or regions where hazardous events are likely to occur, based on the high-risk scenario rules and the scene's initial and target states. Subsequently, to balance vehicle dynamic feasibility with sufficient flexibility and diversity in trajectory generation, we employ a polynomial-based trajectory generation method. This method utilizes the vehicle's initial velocity, initial acceleration, initial position, estimated travel time, and the calculated hazardous target point as core constraints to solve for a unique polynomial trajectory satisfying these conditions. Finally, dense sampling is performed along this polynomial function to simulate a complete, smooth motion trajectory leading to the predefined high-risk state.

\subsubsection{Automated Labeling and Quality Check}

Following the synthesis of a substantial volume of high-risk trajectory data, we implemented meticulous automated labeling and stringent manual quality verification processes.

\noindent\textbf{Automated Feature Extraction and Labeling.} Leveraging the simulator environment, we automatically extract exhaustive kinematic features from each synthesized trajectory. These features include, but are not limited to, high-precision coordinates (x,y)(x,y) at each time step, instantaneous velocity, acceleration, and geometric attributes such as curvature. This structured data provides rich input for subsequent risk analysis and model training.

\noindent\textbf{Trajectory Projection and Human-in-the-Loop Quality Check.} After obtaining the synthesized trajectories from a global (BEV) perspective, we utilize the corresponding camera intrinsic/extrinsic parameters and coordinate transformation matrices to accurately project these BEV trajectories onto the ego-vehicle's first-person view (front-view camera) image plane. This step is crucial for subsequent vision-based risk perception. In conjunction with the aforementioned definitions of high-risk scenarios, we organized a team for manual review to rigorously screen the extensive initial set of synthesized data. The primary objective was to filter out trajectories that were physically implausible (e.g., instantaneous movements violating Newtonian laws), semantically inconsistent with the scene, or could not be clearly and effectively projected onto the camera view due to occlusions, truncations, or other artifacts. Through this meticulous human-in-the-loop quality control process, we ultimately curated approximately 10,000 high-quality, representative samples of high-risk motion data.

\subsubsection{Caption Generation}

Based on the acquired risk category labels, the ego-vehicle's first-person view images, and the detailed kinematic features of the trajectories, we guided GPT-4o~\cite{achiam2023gpt} to organically integrate information from these three dimensions into a unified textual description space. The model was prompted to generate a coherent and accurate natural language "caption." The final generated captions typically encompass three core components: first, a comprehensive description of the current driving scene, including road type, traffic density, weather conditions (if discernible), and key static and dynamic elements; second, an in-depth analysis of the ego-vehicle's planned motion behavior, explaining its potential risk points and possible violations of traffic rules or safety principles; and finally, a clear concluding risk prediction, specifying the likely type of risk (e.g., collision, hard braking, lane departure). Through this structured approach, each high-quality high-risk motion datum was paired with a content-rich and informationally accurate textual description. This process resulted in the construction of a large-scale, multimodal dataset comprising scene-image-trajectory-text-risk label information, laying a solid foundation for subsequent VLM training and risk prediction research.

\begin{figure}[t!]
  \centering
    \includegraphics[width=\linewidth]{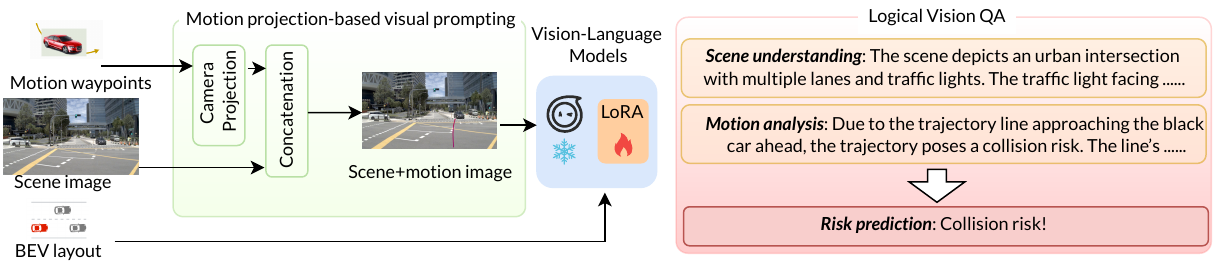}
  \vspace{-0.2cm}
  \caption{
The proposed \ourmethod{} framework. Taking BEV layout, scene images, and motion waypoints as input, \ourmethod{} first visually formats motion using projection-based prompting to overcome modal gap. It then prompts a VLM with combined BEV global context and front-view perspective. \ourmethod{} subsequently infers risk through a 'scene understanding $\rightarrow$ motion analysis $\rightarrow$ risk prediction' chain-of-thought.
  }
  \label{fig: ourmethod} 
\end{figure}

\subsection{DriveMRP-Metric}
\label{sec:ourmetric}
The DriveMRP-Metric is designed to comprehensively evaluate the model's capabilities in driving scene understanding and risk prediction, primarily encompassing the following two aspects:

\noindent\textbf{Cross-modal Scene Understanding Capability.} This capability primarily assesses the model's comprehensive perception and depth of understanding of the current complex driving environment. Specifically, it involves the accurate semantic segmentation and interpretation of images from the driver's perspective, the effective fusion and comprehension of multimodal features including visual, textual semantics, and vehicle kinematics, and the generation of rich and accurate scene description texts based on this understanding. Key metrics for evaluating this capability include: ROUGE-1-F1, ROUGE-2-F1, ROUGE-L-F1~\cite{chin2004rouge}, and BERTScore~\cite{zhang2019bertscore}. These metrics primarily measure the similarity between the generated text and reference texts at the word, phrase, and semantic levels, with higher values indicating stronger scene description and understanding abilities of the model.

\noindent\textbf{High-Risk Motion Prediction Capability.} This capability focuses on evaluating the model's ability to perform precise behavioral analysis of the vehicle's future trajectory and subsequently infer and classify potential high-risk event types, based on a thorough understanding of the driving scene. This requires the model not only to recognize the dynamic characteristics of the trajectory but also to integrate them with the complex scene context to accurately categorize potential hazardous situations. Standard classification evaluation metrics are primarily adopted to assess this capability: Accuracy measures the overall correctness of the model's classifications; Recall reflects the model's ability to identify all instances of a specific high-risk category; and the F1 Score, as the harmonic mean of precision and recall, provides a comprehensive evaluation of classification performance~\cite{goutte2005probabilistic}. Higher values for these metrics indicate superior performance of the model in the identification, reasoning, and classification of high-risk behaviors.

\subsection{\ourmethod{}}
\label{sec:ourmethod}

\subsubsection{Model Architecture}

We utilize Qwen2.5VL-7B~\cite{bai2025qwen2} as our base model, chosen for its exceptional multimodal understanding performance. 
As in Figure~\ref{fig: ourmethod}, our DriveMRP-Agent inputs include BEV layouts, current scene image, and vehicle motion waypoints. To bridge the modal gap between waypoint coordinates and visual data, we implement a motion projection-based visual prompting scheme. This scheme converts motion waypoint information into a visual representation, embedding it into the model's visual processing pipeline for natural fusion with the image modality. Subsequently, our model integrates BEV global context with front-view image to jointly prompt the VLM. Through a logical visual question answering mechanism, the model follows a "Scene Understanding $\rightarrow$ Motion Analysis $\rightarrow$ Risk Prediction" cognitive pathway. The model first understands the driving environment and key traffic participants, then analyzes motion trends and potential interactions, and finally infers and outputs precise predictions of potential driving risks based on integrated information. This structured reasoning enhances predictive accuracy and decision interpretability.

\subsubsection{Training Strategy}

To enhance training efficiency and conserve resources, we employed Low-Rank Adaptation (LoRA)~\cite{hu2022lora} to supervise fine-tuning our base model. Our training strategy draws from Chain-of-Thought (CoT)~\cite{wei2022cot} prompting. 
CoT guides the model to generate intermediate reasoning steps for the final output, improving complex reasoning performance and enhancing decision process interpretability. 
We argue that for driving risk prediction, step-by-step reasoning based on comprehensive scene understanding is pivotal for improving risk identification precision and classification reliability. 
Guided by this, we utilized our \ourdata{} to train the Qwen2.5VL-7B model. Through targeted training, the DriveMRP-Agent acquires CoT-like reasoning, enabling in-depth analysis of complex driving situations and generation of highly accurate, interpretable risk predictions.

\section{Experiments}
\subsection{Datasets}

Due to the scarcity of datasets featuring high-risk motion scenarios, in this work, we constructed \ourdata{}, a synthetic dataset derived from the nuPlan dataset~\cite{caesar2021nuplan}, comprising 10,000 distinct driving scenarios. This dataset was partitioned into training and testing sets at an 8:2 ratio, yielding 8,000 samples for training and 2,000 samples for evaluation. 
Furthermore, to evaluate the real-world generalization capabilities of our model, we curated an independent real-world motion risk dataset consisting of 5,000 driving segments. These segments were collected from actual vehicle operations and exclusively feature genuine high-risk events. The utilization of both \ourdata{} and this real-world dataset allows for a comprehensive assessment of the model's effectiveness in synthetic environments and its generalization in real world. More details about datasets and training details are in Appendix.

\subsection{Main Results}

\setlength{\tabcolsep}{0.4pt}
\renewcommand{\arraystretch}{1.1}
\begin{table}[h]
    \centering
    \small
    \caption{The performance of multiple VLM-based methods on \ourdata{}.}
    \label{tab:metric_on_synthesizedData}
    \begin{tabular}{l|cccc|ccc}
    \toprule
    \multicolumn{1}{l|}{\multirow{2}{*}{\textbf{Method}}} & \multicolumn{4}{c|}{\textbf{Scene Understanding $\uparrow$}} & \multicolumn{3}{c}{\textbf{Motion Risk Prediction $\uparrow$}} \\
    \cmidrule(lr){2-5} \cmidrule(lr){6-8}
    \multicolumn{1}{c|}{} & ROUGE-1-F1  & ROUGE-2-F1  & ROUGE-L-F1  & BERTScore  & Accuracy  & Recall  & F1 score  \\
    \midrule
    EM-VLM4AD-Base~\cite{EM-VLM4AD} & 14.88 & 1.38 & 11.09 & 45.70 & - & - & - \\
    Llava-1.5-7B~\cite{liu2024improvedllava} & 42.67 & 11.44 & 27.23 & 65.18 & 22.34 & 1.72 & 0.85 \\
    InternVL2-8B~\cite{chen2024internvl} & 51.15 & 16.84 & 31.11 & 69.66 & 18.35 & 3.20 & 2.98 \\
    InternVL2.5-8B~\cite{chen2024internvl} & 49.89 & 15.07 & 29.21 & 68.70 & 26.86 & 9.58 &  4.79\\    
    Llama3.2-vision-11B~\cite{grattafiori2024llama3} & 23.50 & 7.07 & 15.48 & 57.10 & 11.32 & 1.12 & 0.83 \\
    Qwen2.5-VL-7B-Instruct\cite{bai2025qwen2} & 48.54 & 15.99 & 30.72 & 68.83 & 27.13 & 13.76 & 6.66 \\ \hline
    \rowcolor{gray!15} 
    \ourmethod{} (ours) & \textbf{69.08} & \textbf{42.23} & \textbf{52.93} & \textbf{81.25} & \textbf{88.03} & \textbf{89.44} & \textbf{89.12}\\
    \bottomrule
    \end{tabular}
    \end{table}

Our empirical evaluations demonstrate the superior capabilities of \ourmethod{} in nuanced scene understanding and accurate risk identification. As presented in Table~\ref{tab:metric_on_synthesizedData}, when compared against contemporary state-of-the-art Vision-Language Models (VLMs) and specialized driving-focused VLMs, \ourmethod{} consistently achieves leading performance across a comprehensive suite of evaluation metrics. This underscores the effectiveness of our approach and its robustness in navigating and interpreting complex, potentially hazardous driving scenarios. Figure 4 further provides qualitative visualizations, illustrating \ourmethod{}'s enhanced ability to discern high-risk trajectories from extreme scenarios and articulate clear, interpretable rationales for its predictions, a capability often lacking in comparative methods.

\begin{figure}[h!]
\centering
\includegraphics[width=1.0\textwidth]{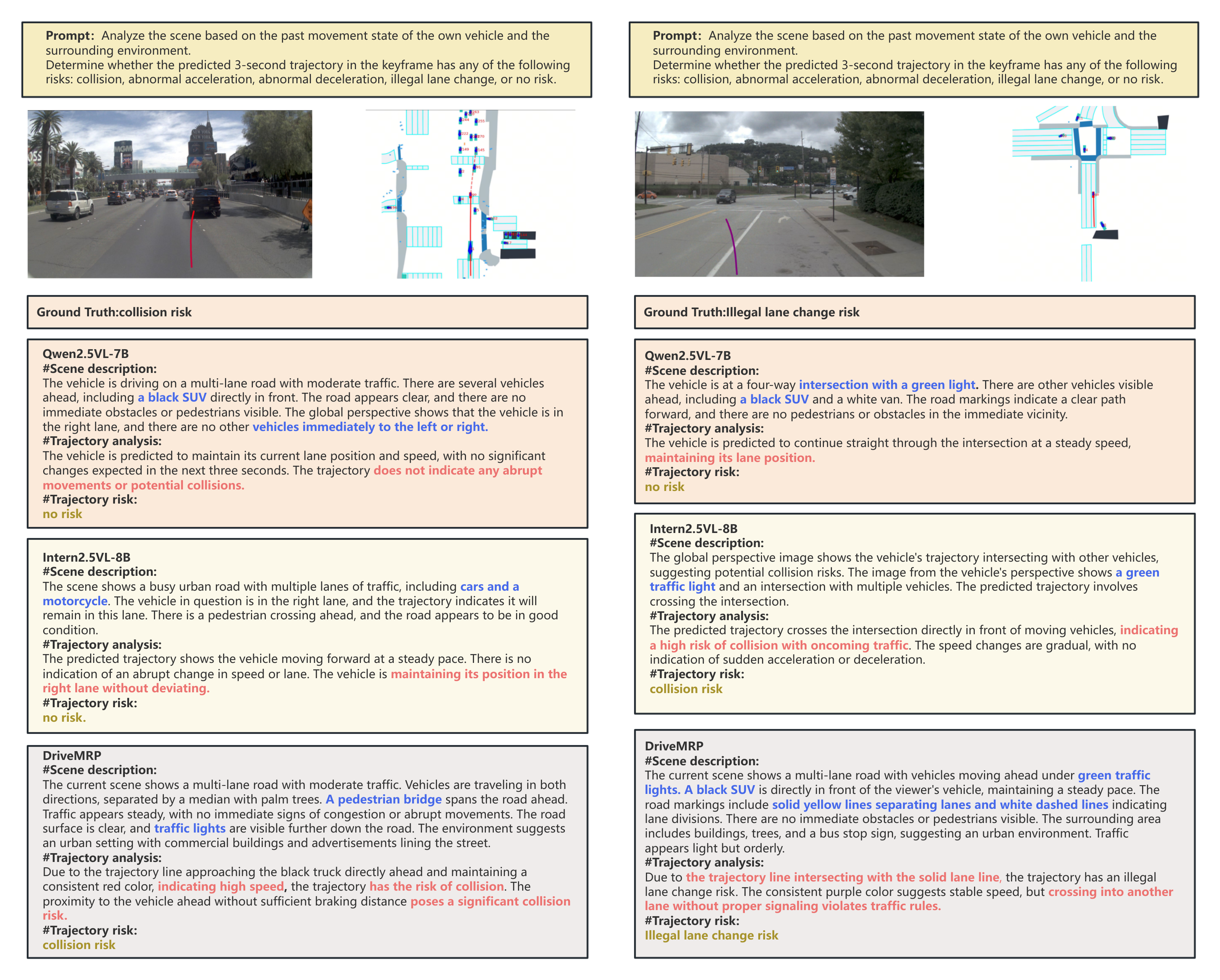}
\vspace{-0,2in}
\caption{This figure shows the performance comparison between our model and other VLM models in scene understanding and risk identification tasks. The highlighted sections demonstrate the key parts of understanding and reasoning. We simultaneously input front-view and bird's-eye view (BEV) images into the model and guide it to output understanding and reasoning results.}
\label{fig:abl_inputs}
\end{figure}

\noindent\textbf{The generalization of \ourmethod{} on unseen real-world risk scenarios.}
To ascertain the generalization prowess of models trained on our synthetic data to unseen real-world conditions, we conducted rigorous testing on our proprietary real-world risk scenario dataset, as detailed in Table~\ref{tab:metric_on_inhouseDataset}. This dataset encompasses a variety of critical events, including collisions, lane encroachments during lane changes, abrupt braking incidents, and sudden accelerations. It is crucial to note that \ourmethod{} was trained solely on the \ourdata{} synthetic dataset and had no prior exposure to these real-world scenarios. Despite this challenging zero-shot transfer setting, \ourmethod{} outperformed generic VLMs by a significant margin of approximately 40\%, achieving a classification accuracy of roughly 70\%. The qualitative results depicted in Figure 5, derived from real-world visual data, further corroborate the effectiveness of our method. These findings validate the powerful generalization capabilities conferred by our innovative visual prompting design and the zero-shot learning potential unlocked through extensive training on diverse synthetic data.

\setlength{\tabcolsep}{0.5pt}
\renewcommand{\arraystretch}{1.1}
\begin{table}[h!]
    \centering
    \small
    \caption{The performance of multiple VLM-based methods on in-house real-world motion datasets.}
    \label{tab:metric_on_inhouseDataset}
    \begin{tabular}{l|cccc|ccc}
    \toprule
    \multicolumn{1}{l|}{\multirow{2}{*}{\textbf{Method}}} & \multicolumn{4}{c|}{\textbf{Scene Understanding} $\uparrow$} & \multicolumn{3}{c}{\textbf{Motion Risk Prediction} $\uparrow$} \\
    \cmidrule(lr){2-5} \cmidrule(lr){6-8}
    \multicolumn{1}{c|}{} & ROUGE-1-F1 & ROUGE-2-F1 & ROUGE-L-F1 & BERTScore & Accuracy & Recall & F1 score \\
    \midrule
    InternVL2-8B~\cite{chen2024internvl}  & 52.42 & 18.19 & 32.44 &  70.72 & 22.75 & 13.65 & 9.55 \\
    InternVL2.5-8B~\cite{chen2024internvl}  & 55.14 & 20.58 & 34.45 & 71.87 & 24.28 & 12.18 & 8.34  \\
    Qwen2.5-VL-7B-Instruct\cite{bai2025qwen2}  & 34.36 & 18.58 & 24.83 & 66.50 & 29.42 & 22.06 & 13.61 \\ \hline
    \rowcolor{gray!15} 
    \ourmethod{} (ours)  & \textbf{62.74} & \textbf{30.82} & \textbf{42.35} & \textbf{76.69} & \textbf{68.50} & \textbf{51.37} & \textbf{56.18} \\
    \bottomrule
    \end{tabular}
    \end{table}

\noindent\textbf{The effectiveness of \ourdata{} for augmenting multiple general-purpose VLMs.}
    The efficacy of our synthetic \ourdata{} dataset is substantiated in Table~\ref{tab:data_aug_on_VLM}. This table details a comparative analysis where various leading VLMs were fine-tuned and evaluated on our dataset. The results consistently indicate that training on \ourdata{} substantially augments the scene understanding and risk identification faculties of these general-purpose VLMs. This highlights the dataset's significant utility and its "plug-and-play" potential in enhancing diverse VLM architectures for driving-related safety applications.

    \setlength{\tabcolsep}{0.5pt}
    \renewcommand{\arraystretch}{1.1}
    \begin{table}[h!]
        \centering
        \small
        \caption{The performance gains of \ourdata{} for multiple general-purpose VLMs.}
        \label{tab:data_aug_on_VLM}
        \begin{tabular}{l|cccc|ccc}
        \toprule
        \multicolumn{1}{l|}{\multirow{2}{*}{\textbf{Method}}} & \multicolumn{4}{c|}{\textbf{Scene Understanding} $\uparrow$} & \multicolumn{3}{c}{\textbf{Motion Risk Prediction} $\uparrow$} \\
        \cmidrule(lr){2-5} \cmidrule(lr){6-8}
        \multicolumn{1}{c|}{} & ROUGE-1-F1 & ROUGE-2-F1 & ROUGE-L-F1 & BERTScore & Accuracy & Recall & F1 score \\
        \midrule
        Llava-1.5-7B~\cite{liu2024improvedllava} & 42.67 & 11.44 & 27.23 & 65.18 & 22.34 & 1.72 & 0.85 \\ \rowcolor{gray!15} 
        + \ourdata{} & \textbf{63.22} & \textbf{34.66} & \textbf{45.57} & \textbf{77.52} & \textbf{59.04} & \textbf{24.11} & \textbf{25.99}  \\
        \midrule
        Llama3.2-vision-11B~\cite{grattafiori2024llama3} & 23.50 & 7.07 & 15.48 & 57.10 & 11.32 & 1.12 & 0.83  \\ \rowcolor{gray!15} 
        + \ourdata{} & \textbf{52.43} & \textbf{33.63} & \textbf{36.47} & \textbf{70.65} & \textbf{56.05} & \textbf{22.04} & 2\textbf{3.03} \\
        \midrule
        Qwen2.5-VL-7B-Instruct\cite{bai2025qwen2} & 48.54 & 15.99 & 30.72 & 68.83 & 27.13 & 13.76 & 6.66  \\ \rowcolor{gray!15} 
        + \ourdata{} & \textbf{69.08} & \textbf{42.23} & \textbf{52.93} & \textbf{81.25} & \textbf{88.03} & \textbf{89.44} & \textbf{89.12}  \\
        \bottomrule
        \end{tabular}
        \end{table}

\subsection{Ablation Studies}

\noindent\textbf{Ablating the effectiveness of multiple inputs in \ourmethod{}.} This subsection presents ablation experiments conducted to dissect the contributions of key components within our \ourmethod{} framework. Specifically, we investigated the importance of incorporating Bird's-Eye View (BEV) information and the relative effectiveness of projecting trajectory data into the image space compared to using raw sequential coordinate information.
As shown in Table~\ref{tab:abl_3inputs}, the results unequivocally demonstrate the substantial benefits of integrating BEV perspective data. Models equipped with BEV input exhibited marked improvements across all evaluated metrics. This finding suggests that providing the model with global contextual information significantly enhances its capacity for comprehensive scene understanding and subsequent inferential reasoning.
Concurrently, we evaluated an alternative approach where the model was trained using raw trajectory coordinate sequences instead of our proposed projection-based visual prompting. This modification led to a pronounced degradation in performance. This outcome lends strong support to our initial hypothesis: Vision-Language Models, by their inherent design, are more adept at interpreting visually represented spatial information, such as lines and paths embedded within an image, than at processing isolated, structured numerical coordinate sequences, which often remain disparate from the rich visual data. Our trajectory projection method effectively bridges this modal gap, enabling the VLM to leverage its powerful visual reasoning capabilities for motion analysis.

\setlength{\tabcolsep}{2.5pt}
\renewcommand{\arraystretch}{1.2}
\begin{table}[h!]
    \centering
    \caption{Ablating three types of inputs in \ourmethod{}: (1) BEV layout, (2) Front-view scene image, and (3) Camera trajectory projection.}
    \vspace{-0.2cm}
    \label{tab:abl_3inputs}
    \begin{tabular}{ccc|cccc|ccc}
    \toprule
     &  &  & \multicolumn{4}{c|}{\textbf{Scene Understanding $\uparrow$}} & \multicolumn{3}{c}{\textbf{Motion Risk Prediction $\uparrow$}} \\
     \cmidrule(lr){4-7} \cmidrule(lr){8-10}
    \multirow{-2}{*}{\textbf{(1)}} & \multirow{-2}{*}{\textbf{(2)}} & \multirow{-2}{*}{\textbf{(3)}} & ROUGE-1-F1 & ROUGE-2-F1 & ROUGE-L-F1 & BERTScore & Accuracy & Recall & F1 score \\
    \midrule
    \Checkmark & \Checkmark & & 68.27 & 41.19 & 52.15 & 80.92 & 85.37 & 84.46 & 83.47   \\
     & \Checkmark  & \Checkmark & 68.75 & 41.85 & 52.54 & 81.19 & 87.50 & 89.01 & 88.22 \\
    \rowcolor{gray!15} 
    \Checkmark & \Checkmark & \Checkmark& \textbf{69.08} & \textbf{42.23} & \textbf{52.93} & \textbf{81.25} & \textbf{88.03} & \textbf{89.44} & \textbf{89.12} \\
    \bottomrule
    \end{tabular}
    \end{table}

\section{Conclusion}
To address risk motion prediction in long-tail autonomous driving scenarios, we introduced \ourdata{}, a synthetic high-risk motion data based on a BEV-based simulation pipeline. We also designed the VLM-agnostic \ourmethod{} framework, using motion-projection visual prompting and chain-of-thought reasoning to enhance risk understanding and interpretability. Trained solely on synthetic data, \ourmethod{} outperformed existing general and driving-specific VLMs, showing strong generalization and paving the way for safer, interpretable autonomous systems.

\noindent\textbf{Limitation, Societal Impact and Future Work.}
One limitation of our work is that the current data synthesis pipeline does not account for sensor-level information, which could improve the realism of the generated motion data. For societal impacts, our work can synthesize high-risk motion data that is difficult to collect, providing training support for developing safer and more robust autonomous driving algorithms. 
In the future, We plan to refine collision classification based on severity levels~\cite{kusano2025comparison} and incorporate spatial reasoning to further improve the performance of motion risk prediction.

{
    \small
    \bibliographystyle{plain}
    \bibliography{11_references}

\begin{thebibliography}{10}

\bibitem{achiam2023gpt}
Josh Achiam, Steven Adler, Sandhini Agarwal, Lama Ahmad, Ilge Akkaya, Florencia~Leoni Aleman, Diogo Almeida, Janko Altenschmidt, Sam Altman, Shyamal Anadkat, et~al.
\newblock Gpt-4 technical report.
\newblock {\em arXiv preprint arXiv:2303.08774}, 2023.

\bibitem{bai2025qwen2}
Shuai Bai, Keqin Chen, Xuejing Liu, Jialin Wang, Wenbin Ge, Sibo Song, Kai Dang, Peng Wang, Shijie Wang, Jun Tang, et~al.
\newblock Qwen2. 5-vl technical report.
\newblock {\em arXiv preprint arXiv:2502.13923}, 2025.

\bibitem{caesar2020_nuscenes}
Holger Caesar, Varun Bankiti, Alex~H Lang, Sourabh Vora, Venice~Erin Liong, Qiang Xu, Anush Krishnan, Yu~Pan, Giancarlo Baldan, and Oscar Beijbom.
\newblock nuscenes: A multimodal dataset for autonomous driving.
\newblock In {\em Proceedings of the IEEE/CVF conference on computer vision and pattern recognition}, pages 11621--11631, 2020.

\bibitem{caesar2021nuplan}
Holger Caesar, Juraj Kabzan, Kok~Seang Tan, Whye~Kit Fong, Eric Wolff, Alex Lang, Luke Fletcher, Oscar Beijbom, and Sammy Omari.
\newblock nuplan: A closed-loop ml-based planning benchmark for autonomous vehicles.
\newblock {\em arXiv preprint arXiv:2106.11810}, 2021.

\bibitem{chen2024autonomous}
Qitong Chen, Dong Zhao, Congzhi Liu, Meng Yang, and Yehui Shi.
\newblock Autonomous vehicles in mixed-autonomy traffic: game theoretic human-like decision making countermeasures.
\newblock {\em Complex Engineering Systems}, 4(4):N--A, 2024.

\bibitem{chen2024internvl}
Zhe Chen, Jiannan Wu, Wenhai Wang, Weijie Su, Guo Chen, Sen Xing, Muyan Zhong, Qinglong Zhang, Xizhou Zhu, Lewei Lu, et~al.
\newblock Internvl: Scaling up vision foundation models and aligning for generic visual-linguistic tasks.
\newblock In {\em Proceedings of the IEEE/CVF conference on computer vision and pattern recognition}, pages 24185--24198, 2024.

\bibitem{chin2004rouge}
Lin Chin-Yew.
\newblock Rouge: A package for automatic evaluation of summaries.
\newblock In {\em Proceedings of the Workshop on Text Summarization Branches Out, 2004}, 2004.

\bibitem{dao2022flashattention}
Tri Dao, Dan Fu, Stefano Ermon, Atri Rudra, and Christopher R{\'e}.
\newblock Flashattention: Fast and memory-efficient exact attention with io-awareness.
\newblock {\em Advances in neural information processing systems}, 35:16344--16359, 2022.

\bibitem{dauner2024navsim}
Daniel Dauner, Marcel Hallgarten, Tianyu Li, Xinshuo Weng, Zhiyu Huang, Zetong Yang, Hongyang Li, Igor Gilitschenski, Boris Ivanovic, Marco Pavone, et~al.
\newblock Navsim: Data-driven non-reactive autonomous vehicle simulation and benchmarking.
\newblock {\em Advances in Neural Information Processing Systems}, 37:28706--28719, 2024.

\bibitem{dosovitskiy2017carla}
Alexey Dosovitskiy, German Ros, Felipe Codevilla, Antonio Lopez, and Vladlen Koltun.
\newblock Carla: An open urban driving simulator.
\newblock In {\em Conference on robot learning}, pages 1--16. PMLR, 2017.

\bibitem{gao2024vista}
Shenyuan Gao, Jiazhi Yang, Li~Chen, Kashyap Chitta, Yihang Qiu, Andreas Geiger, Jun Zhang, and Hongyang Li.
\newblock Vista: A generalizable driving world model with high fidelity and versatile controllability.
\newblock {\em arXiv preprint arXiv:2405.17398}, 2024.

\bibitem{EM-VLM4AD}
Akshay Gopalkrishnan, Ross Greer, and Mohan Trivedi.
\newblock Multi-frame, lightweight \& efficient vision-language models for question answering in autonomous driving.
\newblock {\em arXiv preprint arXiv:2403.19838}, 2024.

\bibitem{goutte2005probabilistic}
Cyril Goutte and Eric Gaussier.
\newblock A probabilistic interpretation of precision, recall and f-score, with implication for evaluation.
\newblock In {\em European conference on information retrieval}, pages 345--359. Springer, 2005.

\bibitem{grattafiori2024llama3}
Aaron Grattafiori, Abhimanyu Dubey, Abhinav Jauhri, Abhinav Pandey, Abhishek Kadian, Ahmad Al-Dahle, Aiesha Letman, Akhil Mathur, Alan Schelten, Alex Vaughan, et~al.
\newblock The llama 3 herd of models.
\newblock {\em arXiv preprint arXiv:2407.21783}, 2024.

\bibitem{han2025DME-Driver}
Wencheng Han, Dongqian Guo, Cheng-Zhong Xu, and Jianbing Shen.
\newblock Dme-driver: Integrating human decision logic and 3d scene perception in autonomous driving.
\newblock In {\em Proceedings of the AAAI Conference on Artificial Intelligence}, volume~39, pages 3347--3355, 2025.

\bibitem{hu2022lora}
Edward~J Hu, Yelong Shen, Phillip Wallis, Zeyuan Allen-Zhu, Yuanzhi Li, Shean Wang, Lu~Wang, Weizhu Chen, et~al.
\newblock Lora: Low-rank adaptation of large language models.
\newblock {\em ICLR}, 1(2):3, 2022.

\bibitem{hu2023uniad}
Yihan Hu, Jiazhi Yang, Li~Chen, Keyu Li, Chonghao Sima, Xizhou Zhu, Siqi Chai, Senyao Du, Tianwei Lin, Wenhai Wang, et~al.
\newblock Planning-oriented autonomous driving.
\newblock In {\em Proceedings of the IEEE/CVF Conference on Computer Vision and Pattern Recognition}, pages 17853--17862, 2023.

\bibitem{huang2024Gen-Drive}
Zhiyu Huang, Xinshuo Weng, Maximilian Igl, Yuxiao Chen, Yulong Cao, Boris Ivanovic, Marco Pavone, and Chen Lv.
\newblock Gen-drive: Enhancing diffusion generative driving policies with reward modeling and reinforcement learning fine-tuning.
\newblock {\em arXiv preprint arXiv:2410.05582}, 2024.

\bibitem{Jiang2023VADVS}
Bo~Jiang, Shaoyu Chen, Qing Xu, Bencheng Liao, Jiajie Chen, Helong Zhou, Qian Zhang, Wenyu Liu, Chang Huang, and Xinggang Wang.
\newblock Vad: Vectorized scene representation for efficient autonomous driving.
\newblock {\em 2023 IEEE/CVF International Conference on Computer Vision (ICCV)}, pages 8306--8316, 2023.

\bibitem{kusano2025comparison}
Kristofer~D Kusano, John~M Scanlon, Yin-Hsiu Chen, Timothy~L McMurry, Tilia Gode, and Trent Victor.
\newblock Comparison of waymo rider-only crash rates by crash type to human benchmarks at 56.7 million miles.
\newblock {\em arXiv preprint arXiv:2505.01515}, 2025.

\bibitem{li2022metadrive}
Quanyi Li, Zhenghao Peng, Lan Feng, Qihang Zhang, Zhenghai Xue, and Bolei Zhou.
\newblock Metadrive: Composing diverse driving scenarios for generalizable reinforcement learning.
\newblock {\em IEEE transactions on pattern analysis and machine intelligence}, 45(3):3461--3475, 2022.

\bibitem{li2024bevformer}
Zhiqi Li, Wenhai Wang, Hongyang Li, Enze Xie, Chonghao Sima, Tong Lu, Qiao Yu, and Jifeng Dai.
\newblock Bevformer: learning bird's-eye-view representation from lidar-camera via spatiotemporal transformers.
\newblock {\em IEEE Transactions on Pattern Analysis and Machine Intelligence}, 2024.

\bibitem{li2024egostatus}
Zhiqi Li, Zhiding Yu, Shiyi Lan, Jiahan Li, Jan Kautz, Tong Lu, and Jose~M Alvarez.
\newblock Is ego status all you need for open-loop end-to-end autonomous driving?
\newblock In {\em Proceedings of the IEEE/CVF Conference on Computer Vision and Pattern Recognition}, pages 14864--14873, 2024.

\bibitem{liao2022maptr}
Bencheng Liao, Shaoyu Chen, Xinggang Wang, Tianheng Cheng, Qian Zhang, Wenyu Liu, and Chang Huang.
\newblock Maptr: Structured modeling and learning for online vectorized hd map construction.
\newblock {\em arXiv preprint arXiv:2208.14437}, 2022.

\bibitem{liu2024improvedllava}
Haotian Liu, Chunyuan Li, Yuheng Li, and Yong~Jae Lee.
\newblock Improved baselines with visual instruction tuning.
\newblock In {\em Proceedings of the IEEE/CVF Conference on Computer Vision and Pattern Recognition}, pages 26296--26306, 2024.

\bibitem{mao2023Gpt-driver}
Jiageng Mao, Yuxi Qian, Junjie Ye, Hang Zhao, and Yue Wang.
\newblock Gpt-driver: Learning to drive with gpt.
\newblock {\em arXiv preprint arXiv:2310.01415}, 2023.

\bibitem{min2024autonomousUnstructured}
Chen Min, Shubin Si, Xu~Wang, Hanzhang Xue, Weizhong Jiang, Yang Liu, Juan Wang, Qingtian Zhu, Qi~Zhu, Lun Luo, et~al.
\newblock Autonomous driving in unstructured environments: How far have we come?
\newblock {\em arXiv preprint arXiv:2410.07701}, 2024.

\bibitem{seo2023safe}
Junwon Seo, Jungwi Mun, and Taekyung Kim.
\newblock Safe navigation in unstructured environments by minimizing uncertainty in control and perception.
\newblock {\em arXiv preprint arXiv:2306.14601}, 2023.

\bibitem{sima2024drivelm}
Chonghao Sima, Katrin Renz, Kashyap Chitta, Li~Chen, Hanxue Zhang, Chengen Xie, Jens Bei{\ss}wenger, Ping Luo, Andreas Geiger, and Hongyang Li.
\newblock Drivelm: Driving with graph visual question answering.
\newblock In {\em European Conference on Computer Vision}, pages 256--274. Springer, 2024.

\bibitem{tian2024drivevlm}
Xiaoyu Tian, Junru Gu, Bailin Li, Yicheng Liu, Yang Wang, Zhiyong Zhao, Kun Zhan, Peng Jia, Xianpeng Lang, and Hang Zhao.
\newblock Drivevlm: The convergence of autonomous driving and large vision-language models.
\newblock {\em arXiv preprint arXiv:2402.12289}, 2024.

\bibitem{wang2024HE-Drive}
Junming Wang, Xingyu Zhang, Zebin Xing, Songen Gu, Xiaoyang Guo, Yang Hu, Ziying Song, Qian Zhang, Xiaoxiao Long, and Wei Yin.
\newblock He-drive: Human-like end-to-end driving with vision language models.
\newblock {\em arXiv preprint arXiv:2410.05051}, 2024.

\bibitem{wang2024omnidrive}
Shihao Wang, Zhiding Yu, Xiaohui Jiang, Shiyi Lan, Min Shi, Nadine Chang, Jan Kautz, Ying Li, and Jose~M Alvarez.
\newblock Omnidrive: A holistic llm-agent framework for autonomous driving with 3d perception, reasoning and planning.
\newblock {\em arXiv preprint arXiv:2405.01533}, 2024.

\bibitem{wang2024Drive-WM}
Yuqi Wang, Jiawei He, Lue Fan, Hongxin Li, Yuntao Chen, and Zhaoxiang Zhang.
\newblock Driving into the future: Multiview visual forecasting and planning with world model for autonomous driving.
\newblock In {\em Proceedings of the IEEE/CVF Conference on Computer Vision and Pattern Recognition}, pages 14749--14759, 2024.

\bibitem{wei2022cot}
Jason Wei, Xuezhi Wang, Dale Schuurmans, Maarten Bosma, Fei Xia, Ed~Chi, Quoc~V Le, Denny Zhou, et~al.
\newblock Chain-of-thought prompting elicits reasoning in large language models.
\newblock {\em Advances in neural information processing systems}, 35:24824--24837, 2022.

\bibitem{zhang2019bertscore}
Tianyi Zhang, Varsha Kishore, Felix Wu, Kilian~Q Weinberger, and Yoav Artzi.
\newblock Bertscore: Evaluating text generation with bert.
\newblock {\em arXiv preprint arXiv:1904.09675}, 2019.

\bibitem{zhang2023perceptionweather}
Yuxiao Zhang, Alexander Carballo, Hanting Yang, and Kazuya Takeda.
\newblock Perception and sensing for autonomous vehicles under adverse weather conditions: A survey.
\newblock {\em ISPRS Journal of Photogrammetry and Remote Sensing}, 196:146--177, 2023.

\bibitem{zheng2024llamafactory}
Yaowei Zheng, Richong Zhang, Junhao Zhang, Yanhan Ye, Zheyan Luo, Zhangchi Feng, and Yongqiang Ma.
\newblock Llamafactory: Unified efficient fine-tuning of 100+ language models.
\newblock In {\em Proceedings of the 62nd Annual Meeting of the Association for Computational Linguistics (Volume 3: System Demonstrations)}, Bangkok, Thailand, 2024. Association for Computational Linguistics.

\bibitem{zhou2024UA-Track}
Lijun Zhou, Tao Tang, Pengkun Hao, Zihang He, Kalok Ho, Shuo Gu, Wenbo Hou, Zhihui Hao, Haiyang Sun, Kun Zhan, et~al.
\newblock Ua-track: Uncertainty-aware end-to-end 3d multi-object tracking.
\newblock {\em arXiv preprint arXiv:2406.02147}, 2024.

\bibitem{ELM}
Yunsong Zhou, Linyan Huang, Qingwen Bu, Jia Zeng, Tianyu Li, Hang Qiu, Hongzi Zhu, Minyi Guo, Yu~Qiao, and Hongyang Li.
\newblock Embodied understanding of driving scenarios.
\newblock In {\em European Conference on Computer Vision}, pages 129--148. Springer, 2024.

\end{thebibliography}
}

\appendix
\section{Technical Details}

\subsection{Training Details}

We select the Qwen-2.5-VL-7B model~\cite{bai2025qwen2} as the foundational model for our \ourmethod{}. Training was conducted exclusively on our synthetic \ourdata{} dataset. We employed the Llama-factory training framework~\cite{zheng2024llamafactory}, a robust and flexible platform for large model training. We use the LoRA~\cite{hu2022lora} framework for fine-tuning the VLMs.
The experiments were conducted using 8 NVIDIA H100 GPUs for distributed training, with FlashAttention~\cite{dao2022flashattention} enabled for acceleration. The learning rate was set to 1.0e-4, the optimizer was AdamW, and a cosine annealing learning rate scheduler was employed.

\end{document}